\documentclass[conference]{IEEEtran}
\IEEEoverridecommandlockouts
\usepackage{cite}
\usepackage{amsmath,amssymb,amsfonts}
\usepackage{algorithmic}
\usepackage{graphicx}
\usepackage{textcomp}
\usepackage{xcolor}
\def\BibTeX{{\rm B\kern-.05em{\sc i\kern-.025em b}\kern-.08em
    T\kern-.1667em\lower.7ex\hbox{E}\kern-.125emX}}
\begin{document}

\title{Embedded Quantum Machine Learning in Embedded Systems: Feasibility, Hybrid Architectures, and Quantum Co-Processors\\
}
\author{
\IEEEauthorblockN{ Somdip Dey}
\IEEEauthorblockA{\textit{School of Engineering and Computing} \\
\textit{Regent College London}\\
London, United Kingdom \\
somdip.dey@rcl.ac.uk, somdipdey.ac@gmail.com}
\and
\IEEEauthorblockN{Syed Muhammad Raza}
\IEEEauthorblockA{\textit{School of Engineering and Computing} \\
\textit{Regent College London}\\
London, United Kingdom \\
syed.raza@rcl.ac.uk}

}

\maketitle

\begin{abstract}
Embedded quantum machine learning (EQML) seeks to bring quantum machine learning (QML) capabilities to resource-constrained edge platforms such as IoT nodes, wearables, drones, and cyber-physical controllers. In 2026, EQML is technically feasible only in limited and highly experimental forms: (i) hybrid workflows where an embedded device performs sensing and classical processing while offloading a narrowly scoped quantum subroutine to a remote quantum processing unit (QPU) or nearby quantum appliance, and (ii) early-stage “embedded QPU” concepts in which a compact quantum co-processor is integrated with classical control hardware. A practical bridge is quantum-inspired machine learning and optimisation on classical embedded processors and FPGAs. This paper analyses feasibility from a circuits-and-systems perspective aligned with the academic community, formalises two implementation pathways, identifies the dominant barriers (latency, data encoding overhead, NISQ noise, tooling mismatch, and energy), and maps them to concrete engineering directions in interface design, control electronics, power management, verification, and security. We also argue that responsible deployment requires adversarial evaluation and governance practices that are increasingly necessary for edge AI systems.
\end{abstract}


\section{Introduction}
Edge and embedded systems increasingly execute machine learning close to sensors to reduce bandwidth, preserve privacy, and enable autonomy. Yet these platforms operate under strict constraints: milliwatt-to-watt power envelopes (indicatively, on the order of ~$\sim$10--1000\,mW for typical sensor nodes and up to a few watts for richer edge SoCs), kilobytes-to-megabytes of memory, limited compute, and deterministic latency requirements (especially in cyber-physical control) \cite{ref24}. QML offers algorithmic primitives—quantum kernels and variational quantum algorithms (VQAs)—that may provide advantages for specific learning and optimisation tasks, but current QPU technologies impose opposite constraints: complex operating environments, non-deterministic runtimes due to sampling, and substantial control overhead.
For the scientific and  engineering community, the practical question is not “Can a fully fledged quantum computer be embedded inside a microcontroller today?” (it cannot), but rather “Which quantum-enabled functions can be integrated into embedded circuits-and-systems architectures, and what co-design is required to make them useful, reliable, and safe?” This paper provides a feasibility-driven view with two implementation pathways: • Pathway 1 (Hybrid): the embedded node runs the classical pipeline and offloads a well-bounded quantum subroutine to a remote QPU (cloud) or to a nearby quantum appliance at the network edge. • Pathway 2 (Embedded QPU co-processor): an on-device quantum module is integrated as a co-processor alongside a classical Microcontroller Unit (MCU) / System-on-a-Chip (SoC) \cite{ref21, ref22} with a low-latency interconnect.
In addition, we discuss quantum-inspired methods as a deployable near-term bridge on classical embedded hardware. Finally, we incorporate security and governance considerations using responsible AI practices (including red teaming) that are increasingly required for trustworthy edge AI deployments \cite{ref1}, and we connect this to open tooling and transparency challenges that arise as quantum and AI capabilities become more democratised \cite{ref2}.

\vspace{0.3em}
\noindent\textbf{Contributions:} This paper (i) formalises two implementable EQML pathways (hybrid offload and embedded QPU/QSoC co-processing) from a circuits-and-systems viewpoint; (ii) identifies dominant feasibility constraints (latency determinism, encoding overhead, NISQ noise, tooling mismatch, energy) and maps them to engineering directions; and (iii) provides an actionable roadmap (including quantum-inspired bridges) with safety, fallback, and governance considerations for trustworthy deployment \cite{ref1,ref2}.
\vspace{0.3em}

\section{BACKGROUND: QML PRIMITIVES AND EMBEDDED CONSTRAINTS}

\subsection{QML Building Blocks}

Most practical quantum machine learning (QML) today targets the noisy intermediate-scale quantum (NISQ) regime~\cite{ref3}. Two families dominate.

\begin{enumerate}
    \item \textbf{Quantum kernel methods:}  
    A parameterised feature map $U(x)$ embeds classical data $x$ into a quantum state; inner products between states define a kernel that can drive classical learners (e.g., support vector machines)~\cite{ref5}. Kernel methods are attractive because the quantum circuit can represent a complex feature space with shallow depth, but they can still be limited by data-encoding cost and shot noise.

    \item \textbf{Variational methods:}  
    A parameterised circuit $U(\boldsymbol{\theta})$ is executed repeatedly; measurement statistics define a cost function that a classical optimiser minimises~\cite{ref4}. Variational quantum classifiers and quantum neural network variants fall into this family. Their practical cost is dominated by repeated circuit execution (shots), hardware noise, and the classical optimisation loop. In NISQ settings, optimisation typically requires repeated circuit executions (often hundreds to thousands of shots per evaluation), amplifying latency and energy costs \cite{ref4}.
\end{enumerate}

\subsection{Embedded Constraints Relevant to QML}

Embedded platforms impose constraints that directly shape QML feasibility:

\begin{enumerate}
    \item \textbf{Latency determinism:}  
    Real-time systems often require bounded response times from microseconds to milliseconds; variable delays can break control stability. Table~\ref{tab:eqml_latency_comparison} summarises how different architectural choices—classical embedded processing, hybrid EQML with remote quantum offload, and on-device QSoC-based EQML—trade off latency, determinism, and feasible task classes. This comparison highlights why hard real-time control loops remain the domain of classical MCUs/SoCs, while EQML is best suited to advisory, background, or optimisation roles under relaxed timing constraints.

    \item \textbf{Energy and thermal limits:}  
    Many embedded nodes run on batteries or energy harvesting, while thermal headroom remains limited; these constraints have long motivated dynamic energy, thermal, and runtime management in embedded and mobile multi-core platforms \cite{ref19,ref20}.

    \item \textbf{Memory and compute limits:}  
    Especially for microcontroller units (MCUs), model storage and intermediate buffers are constrained.

    \item \textbf{Connectivity assumptions:}  
    Some embedded systems are intermittently connected or operate in degraded networks.

    \item \textbf{Safety and reliability:}  
    Decisions must degrade gracefully; stochastic outputs require confidence estimation.
\end{enumerate}

\begin{table}[t]
\caption{Latency, Determinism, and Suitable Tasks for Classical and EQML Architectures}
\label{tab:eqml_latency_comparison}
\centering
\resizebox{\columnwidth}{!}{%
\begin{tabular}{lccc}
\hline
\textbf{Architecture} & \textbf{Typical Latency} & \textbf{Determinism} & \textbf{Suitable Tasks} \\
\hline
Classical MCU / SoC & $\mu$s--ms & High & Control, safety-critical loops \\
Hybrid EQML (remote QPU) & ms--s & Low & Background inference, optimisation \\
QSoC-based EQML (on-device QPU) & ms (est.) & Medium & Advisory inference, tuning \\
\hline
\end{tabular}%
}
\end{table}

A viable embedded quantum machine learning (EQML) design must treat the quantum processing unit (QPU) as a constrained accelerator rather than a general replacement for classical compute. Furthermore, the overall system must remain safe and functional when the quantum component is delayed, unavailable, or noisy.

\section{IMPLEMENTATION PATHWAYS FOR EMBEDDED QML}
In practice, “EQML” covers a spectrum of integration tightness. At one end, a microcontroller simply triggers a remote quantum service; at the other, a quantum module is tightly coupled to the embedded controller over a local bus. Between these extremes lie quantum-inspired methods and quantum-sensor front-ends, which are already deployable in embedded products. 
Fig. 1 depicts the two pathways emphasised in this paper. As summarised in Table~\ref{tab:eqml_latency_comparison} and illustrated in Fig.~1, the degree of quantum–classical integration directly trades off latency, determinism, and feasible application classes.

\begin{figure*}[t]
    \centering
    \includegraphics[width=1.0\columnwidth]{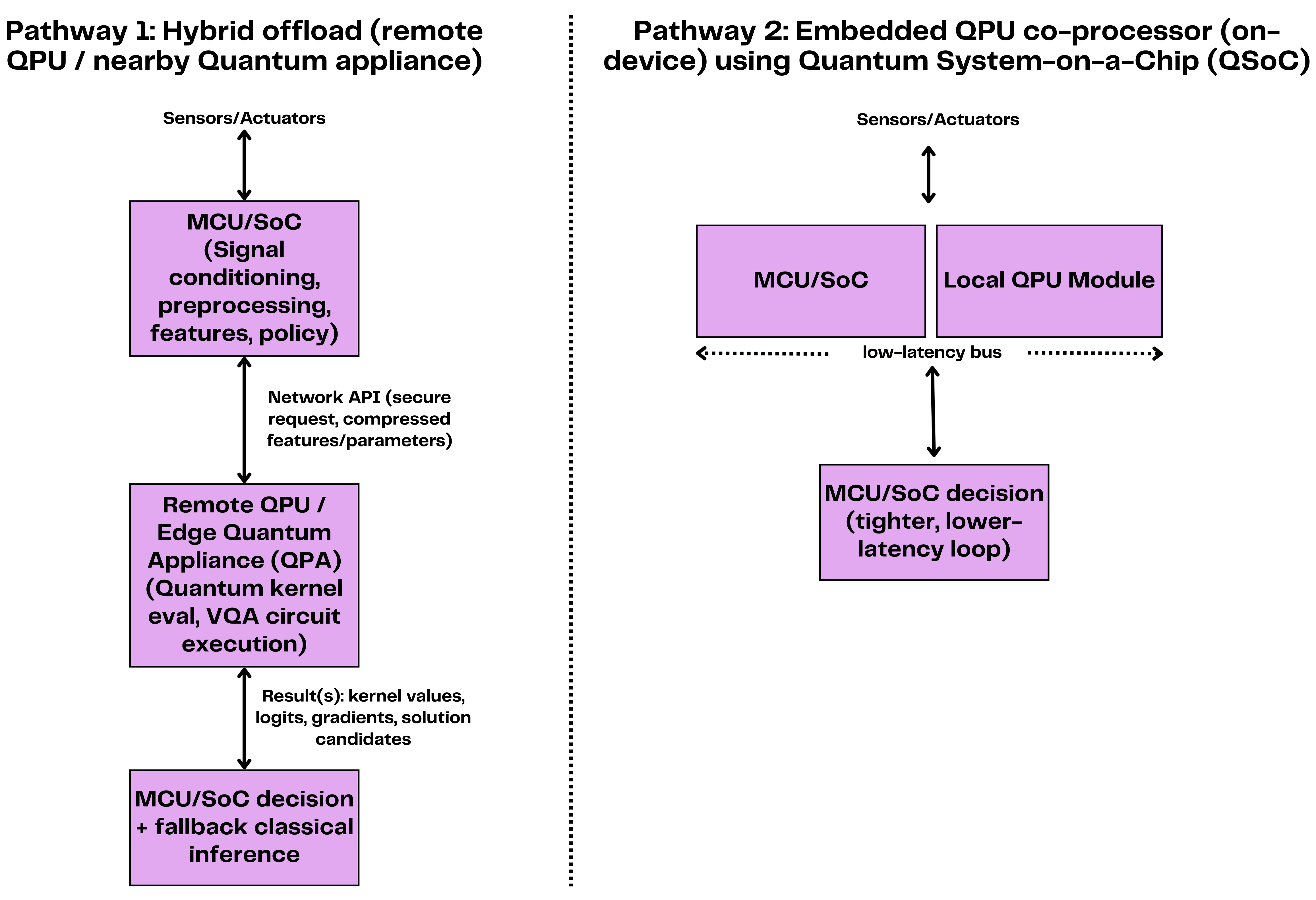}
    \caption{Two EQML implementation pathways. \textbf{Pathway 1 (Hybrid offload):} the embedded MCU/SoC performs sensing and classical preprocessing, then offloads a bounded quantum subroutine to a remote QPU or nearby quantum appliance (QPA), returning kernel values/logits/gradients for fusion with a classical fallback. \textbf{Pathway 2 (On-device QSoC):} a local QPU module is coupled to the MCU/SoC via a low-latency interconnect to enable tighter quantum--classical loops without network dependence.}
    \label{fig:placeholder}
\end{figure*}

\subsection{Pathway~1: Hybrid Embedded--Quantum Systems}
\label{subsec:pathway1}

\subsubsection{Partitioning and Dataflow}

In Pathway~1, the embedded node performs sensing, classical preprocessing, feature extraction, and system control, while the quantum processing unit (QPU) executes a narrowly defined quantum subroutine. The quantum backend may be a cloud QPU accessed via an application programming interface (API), or a QPU placed closer to the edge (e.g., on premises within an industrial network) to reduce round-trip delay.

Typical offloaded tasks include:
\begin{itemize}
    \item quantum kernel evaluations for compact classifiers;
    \item evaluation of shallow variational circuits as part of a classical optimisation loop;
    \item quantum-assisted optimisation steps for discrete decision problems.
\end{itemize}

The embedded node fuses quantum outputs into its decision logic (e.g., anomaly scores, classification logits, or candidate solutions) while maintaining a classical fallback path.

\subsubsection{Feasible Application Classes}

Hybrid embedded quantum machine learning (EQML) is most plausible when quantum calls are infrequent, batched, or asynchronous.

\paragraph{ IoT anomaly detection and security}
A systematic review highlights QML opportunities for IoT security while emphasising embedded resource constraints and the need for lightweight feature engineering~\cite{ref6}. A concrete example is federated quantum kernel learning for multivariate IoT time series, where edge nodes compute compressed kernel statistics using parameterised quantum circuits and share only summaries with a server that trains a decision function~\cite{ref7}. This structure aligns with embedded constraints: local preprocessing is performed on-device, communication payloads remain small, and the quantum component is used selectively where kernel expressiveness may help.

\paragraph{ Embedded robotics and autonomy}
Quantum kernels or small optimisation subroutines could be queried to refine perception or planning, but practical use is limited to non-critical loops (e.g., periodic map refinement or background model updates). High-rate stabilisation and safety-critical control must not depend on remote QPU calls. It is important to clarify that EQML is not suitable for hard real-time control loops requiring microsecond-level determinism, such as flight stabilisation, braking systems, or safety-critical motor control. In such systems, quantum components—whether remote (Pathway~1) or local (Pathway~2)—must remain outside the critical control path, serving only as advisory or background optimisation modules whose outputs are fused by classical controllers under deterministic scheduling constraints.

\paragraph{ Authentication and trust services}
Hybrid quantum services can provide high-quality randomness or challenge--response primitives within a security architecture. These services must be integrated with classical cryptographic controls and robust network security mechanisms.

\subsubsection{Circuits-and-Systems Implications}

For embedded systems designers, Pathway~1 is primarily an interface and co-design problem. It benefits from workflow orchestration principles developed in hybrid quantum--classical computing, where classical systems schedule QPU jobs and fuse results into a larger pipeline. This orchestration style is increasingly discussed in quantum--HPC integration literature~\cite{ref17,ref18}.

Key implications include:
\begin{itemize}
    \item \textbf{Feature compression and encoding-aware preprocessing:} Reducing feature dimensionality can lower quantum encoding depth and improve latency.
    \item \textbf{Secure and efficient communications:} The embedded--QPU boundary must ensure confidentiality and integrity, while supporting caching, batching, and timeout control.
    \item \textbf{Deterministic integration:} Quantum results must be treated as optional accelerations; the embedded controller must remain correct when results are delayed or unavailable.
\end{itemize}

\subsubsection{Limitations}

Hybrid EQML is constrained by:
\begin{itemize}
    \item \textbf{Communication latency and queueing:} Round-trip delays and cloud job queues are incompatible with hard real-time control loops such as flight control, braking, and high-frequency motor control. In practice, network round-trip plus queuing can easily push end-to-end inference indicatively into the order of 10–1000 ms regime (or higher under shared access), which is incompatible with microsecond--millisecond control deadlines.
    \item \textbf{Encoding overhead:} Mapping high-dimensional sensor data into quantum states can dominate execution cost.
    \item \textbf{Tooling mismatch:} QML frameworks are Python-centric, whereas embedded systems rely on C/C++ and real-time operating systems (RTOS).
\end{itemize}

These limitations motivate quantum-inspired bridges and the longer-term pursuit of embedded quantum co-processors implemented as quantum systems-on-chip (QSoCs).

\subsection{Quantum-Inspired Methods on Embedded Hardware}
\label{subsec:quantum_inspired}

Quantum-inspired methods run on classical hardware but exploit structures motivated by quantum states or quantum optimisation. They are attractive for embedded systems because they can deliver practical gains without requiring a QPU.

\subsubsection{Tensor-Network Learning and Compression}

Tensor networks, originally developed for quantum many-body physics, provide an effective machine learning paradigm with natural compression mechanisms~\cite{ref8}. Their structure can significantly reduce parameter counts and memory footprints while preserving expressiveness, making them attractive for embedded inference. Recent surveys discuss their applicability in industrial contexts and highlight scalability limits~\cite{ref9}.

\subsubsection{Quantum-Inspired Optimisation Heuristics}

Annealing- and QAOA-inspired heuristics implemented classically—often accelerated using FPGAs or GPUs—can address combinatorial optimisation problems such as scheduling and resource allocation. In embedded systems, these heuristics can be applied to planning or control subproblems where exact optimisation is infeasible.

Quantum-inspired methods also complement hybrid QML by compressing or selecting features prior to quantum encoding, thereby reducing circuit depth and communication payloads.

\subsection{Embedded Devices as Quantum Sensor Front-Ends and Simulation Nodes}
\label{subsec:quantum_sensing}

Even when QML computation remains remote, embedded platforms can provide quantum-native inputs. Nitrogen-vacancy (NV) centre diamond devices and other quantum sensors can be integrated as compact sensing modules whose outputs enhance classical and potentially quantum-assisted inference. In this framing, the embedded device hosts quantum sensing rather than universal quantum computing, followed by local classical ML or remote hybrid QML.

For circuits-and-systems designers, this raises familiar challenges—analogue front-ends, noise filtering, digitisation, calibration, and power management—now centred on quantum-grade sensing elements.

A complementary role is local simulation and verification of small quantum circuits. Embedded GPUs or FPGAs can accelerate simulation for small qubit counts, enabling on-device testing of encoding circuits, verification of compiled kernels before submission to a remote QPU, and deterministic fallbacks when connectivity is unavailable. While simulation does not provide quantum advantage, it reduces integration risk and supports safety-critical validation.

\subsection{Pathway~2: Embedded Quantum Co-Processors}
\label{subsec:pathway2}

\subsubsection{Concept}

Pathway~2 integrates a local QPU alongside the classical processor within the embedded platform, analogous to an accelerator block in a heterogeneous system-on-chip. A low-latency interconnect enables tighter quantum--classical loops, allowing circuits to be dispatched without network dependency. If realised, this pathway could support more frequent quantum calls within constrained latency budgets.

\subsubsection{Candidate Hardware Directions and Evidence}

Most QPU modalities remain incompatible with embedded form factors; however, several directions show progress toward modular accelerators:

\begin{itemize}
    \item \textbf{NV-centre diamond devices:} Room-temperature diamond quantum accelerators have been demonstrated in rack-scale systems, with ongoing efforts toward mobile quantum computers~\cite{ref10,ref11,ref12}.
    \item \textbf{Photonic integrated quantum hardware:} Photonics supports chip-scale, room-temperature operation. Modular photonic architectures and on-chip generation of error-resistant qubits have been reported~\cite{ref13,ref14}.
    \item \textbf{Chip-integrated optics for trapped ions:} Although trapped-ion systems require vacuum, photonic integration can reduce size and improve manufacturability, as demonstrated by recent industry collaborations~\cite{ref15}.
\end{itemize}

\subsubsection{Circuits-and-Systems Implications}

Pathway~2 reframes EQML as a classical control and integration problem involving low-power mixed-signal electronics, packaging and shielding, calibration, self-test, and deterministic scheduling. Hardware--software co-design is essential, including compilers and runtimes that manage calibration constraints and provide predictable execution semantics~\cite{ref16}.

\subsubsection{Realistic Near-Term Roles}

Near-term embedded QPUs are likely to be small (tens of qubits) and highly specialised. Practical early roles include quantum sensing modules, quantum random number generation, and narrow accelerators for shallow circuits. Broader QML workloads will remain constrained until fault-tolerant qubits and energy-efficient control stacks mature.

\section{Key Challenges and Engineering Directions}
\label{sec:challenges}

\subsection{Hardware Environment Mismatch}
\label{subsec:hardware_mismatch}

Quantum hardware typically requires isolation from thermal and electromagnetic noise, whereas embedded platforms are constrained, mobile, and cost-driven. This fundamental mismatch limits direct integration. Promising research directions include higher-temperature qubit modalities (e.g., diamond and photonics), modular packaging approaches that bring QPUs closer to the edge, and co-designed control application-specific integrated circuits (ASICs) that reduce wiring complexity, radio-frequency (RF) overhead, and calibration burden.

\subsection{NISQ Noise, Limited Qubits, and Reliability}
\label{subsec:nisq_reliability}

Noisy intermediate-scale quantum (NISQ) devices suffer from decoherence, gate errors, and measurement noise, and QML workloads often require repeated sampling. Embedded feasibility therefore depends on shallow circuits, effective error mitigation, and robust uncertainty handling. For safety-critical systems, architectures must incorporate classical fallbacks, confidence estimation, and conservative decision thresholds, rather than treating quantum outputs as deterministic signals.

\subsection{Latency: Network and On-Device Scheduling}
\label{subsec:latency}

In Pathway~1, latency includes both network delay and cloud queueing effects; in Pathway~2, latency arises from on-device scheduling, calibration requirements, cooldown periods, and duty-cycling constraints. Mitigation strategies include restricting quantum calls to background tasks (e.g., periodic retraining or parameter tuning), deploying QPUs closer to the edge to reduce round-trip delay, and designing task graphs that allow the system to progress safely even when quantum results are delayed or unavailable.

\subsection{Classical-to-Quantum Data Encoding}
\label{subsec:encoding}

Encoding classical data into quantum states is frequently the dominant cost in QML pipelines. Embedded feasibility improves when feature maps are low-depth and matched to the sensor modality, or when classical preprocessing reduces dimensionality prior to encoding. Promising directions include structured encodings (e.g., angle encoding for low-dimensional features), feature selection, and joint learning of classical front-ends with quantum feature maps to minimise circuit depth and execution cost.

\subsection{Toolchain Mismatch, Verification, and Portability}
\label{subsec:toolchains}

QML toolchains are dominated by Python-based frameworks (e.g., Qiskit, PennyLane, and Cirq), whereas embedded engineering relies on lightweight runtimes, C/C++ interfaces, real-time operating systems (RTOS), and deterministic testing. This mismatch complicates deployment and verification. Embedded systems often require static analysis, worst-case execution-time reasoning, and reproducible builds, while QML stacks evolve rapidly and frequently depend on cloud services.

Academia-aligned research directions include embedded-friendly intermediate representations and runtime libraries for quantum kernels and variational algorithms, along with driver-level interfaces for local QPUs. Co-simulation and debugging infrastructures are essential to validate mixed quantum--classical designs and to support reproducible benchmarking~\cite{ref16}.

\subsection{Energy Consumption}
\label{subsec:energy}

Embedded power budgets are often incompatible with current QPU control stacks. Even room-temperature quantum accelerators typically target rack-scale or edge-server power envelopes. Progress requires energy-aware control electronics, duty cycling, and burst-oriented usage models in which quantum calls are rare and carefully scheduled. Continued hardware innovation to reduce readout and control energy, combined with system-level power management and thermal design, will be central to making embedded quantum co-processors viable.

\subsection{Safety, Security, and Robustness}
\label{subsec:safety}

Quantum-augmented edge AI introduces new failure and attack surfaces. Adversarial inputs may exploit encoding choices, stochastic outputs can be amplified by downstream control logic, and hybrid cloud interfaces increase exposure. Responsible AI practices—particularly systematic adversarial evaluation (``red teaming'')—provide a practical framework for testing quantum-augmented pipelines under worst-case conditions, including latency faults, data poisoning, and model inversion attacks~\cite{ref1}.

Open and transparent evaluation is also important as quantum and AI capabilities become more accessible. While open models and tooling improve reproducibility, they also alter threat models and necessitate stronger governance and security controls~\cite{ref2}.

\section{Roadmap and Research Opportunities for Academia and Engineering}
\label{sec:roadmap}

Embedded quantum machine learning (EQML) is best understood as a staged technology roadmap rather than a single deployment target:

\begin{itemize}
    \item \textbf{Now (2026):} Hybrid quantum offload is feasible for asynchronous and non-critical tasks. Quantum-inspired methods are practical on microcontrollers (MCUs), field-programmable gate arrays (FPGAs), and embedded GPUs. Embedded systems can integrate quantum sensors or quantum randomness components, and local accelerators can support quantum circuit simulation and verification to enable robust integration testing.

    \item \textbf{Near term (3--5 years):} More ruggedised and edge-deployable quantum appliances may reduce latency for industrial settings. Embedded software stacks and toolchains may mature toward lighter runtimes, improved interfaces, and better integration with real-time operating systems.

    \item \textbf{Midterm (5--10 years):} Prototype embedded quantum co-processors implemented as quantum systems-on-chip (QSoCs) may emerge for niche tasks such as sensing, randomness generation, and narrow acceleration of shallow circuits. Tighter quantum--classical integration frameworks will become increasingly important.

    \item \textbf{Longer term (10+ years):} If fault-tolerant qubits and energy-efficient control electronics mature, more general EQML co-processors and real-time quantum-assisted inference may become feasible within embedded latency and power budgets.
\end{itemize}

Within this roadmap, several research opportunities are particularly well aligned with academic and engineering communities:

\begin{enumerate}
    \item \textbf{Control and readout ASICs and mixed-signal front-ends:}  
    Design of low-power, highly integrated control electronics that reduce QPU overhead related to wiring, RF generation, digitisation, and calibration, enabling compact and modular quantum accelerators.

    \item \textbf{Deterministic scheduling and real-time APIs:}  
    Development of real-time interfaces for QPU invocation, including timeout-aware execution, graceful degradation, and uncertainty-aware fusion of quantum and classical outputs.

    \item \textbf{Hardware--software co-design for encoding-aware pipelines:}  
    Joint optimisation of sensor feature extraction, classical preprocessing, and quantum data encoding to minimise circuit depth, latency, and power consumption.

    \item \textbf{Design abstractions and co-simulation workflows:}  
    Creation of hardware description languages (HDLs), intermediate representations, and co-simulation environments for tightly integrated quantum--classical embedded systems, supporting verification and reproducible experimentation~\cite{ref16}.

    \item \textbf{Security-by-design for hybrid quantum systems:}  
    Integration of security and robustness considerations from the outset, including adversarial testing (``red teaming''), open evaluation practices, and governance mechanisms for hybrid quantum services and embedded QPU modules~\cite{ref1,ref2}.
\end{enumerate}

\subsection{A Speculative Future Trajectory for EQML}
Beyond conventional edge inference, the longer-term evolution from embedded machine learning (TinyML/edge AI) toward embedded quantum machine learning (EQML) may enable new human--computer interaction paradigms. One emerging direction is ``neural coding''---interpreted here as translating human intent (e.g., thoughts-to-code or intent-to-implementation) into executable software artefacts---which has been discussed as a plausible successor to ``vibe coding'' in the AI era \cite{ref23, ref25, ref26}. While current EQML is not positioned for hard real-time control loops (Table~I), future QSoC-based EQML could contribute as an on-device accelerator for complex optimisation and representation-learning subroutines that underpin such intent-driven software engineering workflows. Conceptually, this would require tightly integrated quantum--classical toolchains, deterministic orchestration, and verification support (cf.\ Pathway~2 and quantum--classical co-design considerations), together with robust governance and adversarial evaluation practices as capabilities become more accessible \cite{ref1,ref2}. We emphasise that this vision remains speculative: it does not imply near-term feasibility of full thoughts-to-code compilation on embedded platforms, but it motivates an additional research trajectory in which EQML advances are co-designed with sensing, interfaces, and trustworthy systems engineering to support richer intent-aware computing at the edge.

\section{Conclusion}
\label{sec:conclusion}

Embedded quantum machine learning (EQML) is feasible today primarily through hybrid architectures that offload narrowly scoped quantum subroutines, as well as through quantum-inspired algorithms that already run on classical embedded hardware. A second pathway—embedded quantum co-processors—remains experimental, but is supported by recent progress in room-temperature and chip-integrated quantum technologies and by emerging work on tightly coupled quantum--classical system design.

Across both pathways, the dominant barriers include hardware environment constraints, NISQ noise and limited qubit counts, classical-to-quantum encoding overhead, latency, tooling maturity, and energy budgets. Overcoming these challenges requires an interdisciplinary circuits-and-systems effort encompassing control and readout electronics, interconnects, packaging, power management, and system-level verification.

Finally, as quantum-augmented edge artificial intelligence moves from conceptual exploration to real-world deployment, security, safety, and governance practices—such as adversarial evaluation (``red teaming'') and transparent benchmarking—must be embedded into the engineering lifecycle to ensure robust, trustworthy, and responsible systems.

\end{document}